# EMOTION ANALYSIS OF SONGS BASED ON LYRICAL AND AUDIO FEATURES

Adit Jamdar[1], Jessica Abraham[2], Karishma Khanna[3] and Rahul Dubey[4]

[1]Department of Computer Engineering and Technology, Veermata Jijabai Technological Institute, Mumbai, India
[2]Department of Computer Engineering and Technology, Veermata Jijabai Technological Institute, Mumbai, India
[3]Department of Computer Engineering and Technology, Veermata Jijabai Technological Institute, Mumbai, India
[4]Department of Computer Engineering and Technology, Veermata Jijabai Technological Institute, Mumbai, India

*ABSTRACT*

*In this paper, a method is proposed to detect the emotion of a song based on its lyrical and audio features. Lyrical features are generated by segmentation of lyrics during the process of data extraction. ANEW and WordNet knowledge is then incorporated to compute Valence and Arousal values. In addition to this, linguistic association rules are applied to ensure that the issue of ambiguity is properly addressed. Audio features are used to supplement the lyrical ones and include attributes like energy, tempo, and danceability. These features are extracted from The Echo Nest, a widely used music intelligence platform. Construction of training and test sets is done on the basis of social tags extracted from the last.fm website. The classification is done by applying feature weighting and stepwise threshold reduction on the k-Nearest Neighbors algorithm to provide fuzziness in the classification.*

## 1. INTRODUCTION

Music is said to be the language of emotions and the activity of listening to music is indeed a part of everyday life. If questioned about the song one would want to hear at any particular moment, one would surely pick a song that would be relevant to his mood. Reliable emotion based classification systems are required to facilitate this. The task of music information retrieval is an intriguing one. However, research in the field of emotion based music classification has not yielded the best results, especially systems that make use of lyric based analysis. Musical aspects definitely play an important role in deciding the emotion of a song. Even so, most people are able to connect with the words of a song better than its musical features. In most cases, the words of the song are what truly express the emotions associated with the music, while the musical aspects are generally made to revolve around the lyrical theme.

In this paper, Section 2 will first explore related works on Support Vector Regression, Thayers co-ordinate model, fuzzy clustering, usage of affective lexicons and social tags, and the tf*idf metric. Section 3 introduces the feature set used for classification, as well as the feature scaling technique. The feature set includes lyrical features like arousal and valence, which were extracted from a dictionary database comprising of three dictionaries that cover virtually all meaningful words that are present in the song lyrics. Association rules based on the POS tags of each lyric line have been developed in order to improve the accuracy of arousal and valence values. These features are supplemented with audio features like beats per minute,danceability, loudness, energy and mode. Section 4 is devoted to the construction and pre-processing of the training dataset. Categories of songs created using crowd-sourced tags from the popular music discovery website Last.fm served as the training dataset for the classifier. Section 5 describes the k-Nearest Neighbors algorithm and the Euclidean distance metric that has been used for classification. It also lists the process followed while classifying a new song. Section 6 includes various experiments such as feature weighting analysis, the effect of number of neighbors considered, the contribution of step-wise threshold reduction to accuracy and fuzziness, expanding ANEW with WordNet and dealing with context and ambiguity.Section 7 contains the results for this classification on four test sets, and will conclude this paper.

## 2. RELATED WORK

### 2.1. Music analysis based on audio features

Many researchers have focused on music retrieval by focusing on genre classification using audio features and meta-data about the song [1] as well as low level feature analysis (such as pitch, tempo or rhythm) [2], while music psychologists have been interested in studying how music communicates emotion. A lot of research has been done on music analysis based on audio features [3], [4] but lyrical analysis has not been explored much.

Other researchers have investigated the influence of musical factors like loudness and tonality on the perceived emotional expression [5], [6]. They analyzed these factors by measuring the psychological relation between each musical factor and what emotion it evokes. According to [5],Juslin and Sloboda investigated the utilization of acoustic factors in the communication of music emotions by performers and listeners and measured the correlation between emotional expressions (such as anger, sadness and happiness) and acoustic cues (such as tempo, spectrum and articulation).

### 2.2.Music Information Retrieval from Lyrics

This paper builds on the assertion that text classification cannot be directly applied to classify the lyrics of a song, since identifying their meaning is much more subtle than a normal document (e.g. news article, movie reviews, stories), which can be captured by applying linguistic principles on lyrics. These linguistic principles are further elaborated in section 3.2.1. A survey [7] of text classification techniques, feature selection and performance evaluation was examined which helped to review and compare algorithms and examine their tradeoffs. N-grams, being one of the most popular features in text classification that are used in tasks like sentiment analysis of movie reviews [8], were also considered for this purpose. But n-grams fail to capture the context and poetic expressions of lyrics. Features like tf*idf turn out to be really useful for document classification, especially documents which are long and have distinctive vocabulary for each document category. It has been successfully employed in categorization of Web documents [9]. They have been used for lyrics classification as well [10], however they are known to not work well with documents of short length and large vocabulary. It also makes no use of semantic similarities between words. Therefore, it fails to mine the context or emotion of the document. Also, emotion has various dimensions and is not confined to the usual positive-negative areas.

Emotions are best conveyed by the lyrics of a song, a fact that is not capitalized upon very well by the existing research. Therefore, this research explores the lyrical aspects of the song to analyze its emotion.

## 2.3. Emotion Model

Emotions can be illustrated in two ways: core affect and prototypical emotional episode [11]. Core affect refers to consciously accessible elemental processes of pleasure and activation (valence and arousal), has many causes, and is always present. Its structure involves bipolar dimensions (valence and arousal). Prototypical emotional episode refers to a complex process that unfolds over time, involves casually connected sub-events (antecedent, appraisal, and self-categorization). Even though related, core affect and prototypical emotional episodes are different concepts. Core affective feelings vary in intensity, and a person is always in some state of core affect [11]. It can be seen as unrelated to prototypical emotional episodes, such as feeling miserable from infection, joy or sadness from listening to a song, feeling excited while reading a mystery novel, etc. As songs do not involve any sub-events as present in prototypical emotional episodes, we consider core affect for the illustration of emotions. Another well-known model, the PANA model is defined by two axes reflecting two basic behavioral systems. Positive Activation (PA) ranges from mood terms like active, elated, and excited to drowsy, dull, and sluggish. The other axis Negative Activation (NA) covers distressed, fearful, nervous at one end and calm, at rest, and relaxed at the other. Analysis of PANA model shows that it is a mere 45 degree rotation of Russell's circumplex model [12]. A more complex model, Putchik's wheel of emotions is often used for sentiment analysis. It extends the basic emotions of Russell's circumplex model and adds more complex emotions like contempt, submission [13]. However, such detailed emotion categories seem redundant when tagging a song for a listener, hence we are primarily focused on the circumplex model which is widely used in lyric classification.

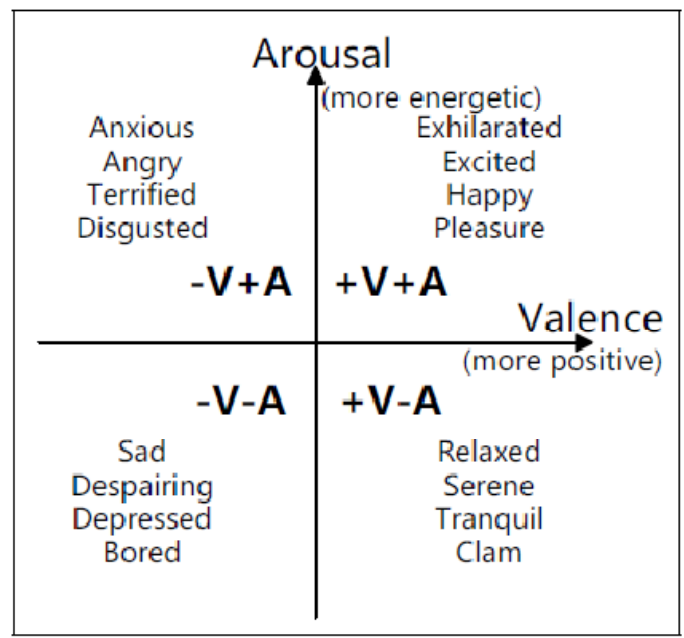

Figure 1: Rusell's Model of Mood

Several papers make use of Russell's model, a two-dimensional model to rate emotions [3], [14], [15], [16]. This approach relies on assigning emotions based on their locations on the two-dimensional co-ordinate system. This model defines two dimensions, viz. Valence and Arousal. Valence is a subjective feeling of pleasantness or unpleasantness; arousal is a subjective state of feeling activated or deactivated. As shown in Figure 1, the two-dimensional emotion plane can be divided into four quadrants with sixteen emotion adjectives placed over them.

## 2.3. Use of Social Tags

We know that music is created by humans for human consumption and hence it encapsulates a lot of contextual knowledge. Computers cannot bring the same level of contextual knowledge, often none at all. Therefore, many problems in Music Information Retrieval are intractable. To expand the pool of contextual knowledge about music for computers, researchers make use of a set of social tags that humans apply to music.

Social tags come into being as a result of many individuals tagging the songs with text annotations, typically to organize their personal content. These tags are then combined with tags created by other individuals to form a collective body of social tags. Last.fm is one such commercial website that allows a user to apply social tags to tracks, albums, artists and playlists. Unfortunately, there is also a great deal of irrelevant information in the tags.Moreover,mood is very subjective hence sometimes songs end up having contradictory tags, which makes it difficult to construct a dataset using these social tags. Nonetheless, they are a source of human-generated contextual knowledge and are widely used in MIR systems. [14], [15], [17], [18].We have constructed our dataset using tags from Last.fm after carrying out filtering for relevant tags.

## 3. FEATURES USED

### 3.1. Feature Selection

On the lines of Russell's Model, the aim is to have features that can classify songs along two axes:

- Positive emotions versus Negative emotions
- High energy versus Low energy

Valence and mode are the two features that are used to represent the first(X) axis. Valence is the key lyrical feature that is used to denote positivity or negativity of emotions. Its values are drawn from the ANEW dictionary and represent an average person's perceptions of the positivity or negativity of a word, and thus are very useful for lyric analysis.Musical Mode is a theoretical concept in music involving note and scale selection, and which plays a large role in the overall positivity or negativity of a song. Listeners perceive songs that are in different modes to be either happy or sad.

Beats per minute, loudness, energy and arousal are the features that represent the second(Y) axis. Arousal is the only lyrical feature on this axis, and draws from values in the ANEW dictionary to determine how much excitement an average listener would perceive from the lyrics. The other features are all drawn from the Echonest music analysis servers. Beats per minute is a good representation of the speed of a song, which can be directly co-related with the energy felt by a listener. Loudness of a song generally reflects the types of instruments used. Heavy distorted guitars and aggressive drum beats tend to be associated with high energy tracks, while softer, acoustic guitars or keyboards tend to be comparatively mellow.

International Journal of Artificial Intelligence & Applications (IJAIA) Vol. 6, No. 3, May 2015

Another feature that has been selected is Danceability. Danceability specifically measures how well a song fits into the first quadrant, that is, positive songs with high energy.

### 3.2. Features

#### 3.2.1. Arousal and Valence

Arousal and Valence (AV) are the primary lyric-based features used by the system. These features are based on the concept of Thayer's Model of Emotion [16]. The Arousal feature provides a measure of the intensity of the emotion that is conveyed by the song. It is measured on a scale of 0 to 10, with the value proportional to the intensity of emotion. The Valence of a song provides an indication of the type of emotion (such as happy or sad) associated with the song. It is also measured on a scale of 0 to 10, with lower values representing negative emotions and higher values representing positive emotions.

AV features for a song are calculated by taking the weighted mean of the AV values of each sentence in the song lyrics. This is shown in Equation 1.

$$Valence_{song} = \sum \frac{Valence_{sentence} \times Weight_{sentence}}{Weight_{total}}$$

$$Arousal_{song} = \sum \frac{Arousal_{sentence} \times Weight_{sentence}}{Weight_{total}}$$

$$where, Weight_{total} = Total\ weight\ of\ all\ sentences$$

Equation 1: Valence and Arousal

The calculation for each sentence is performed by considering the individual AV values of words in the sentence. These values are extracted from a dictionary database consisting of 3 dictionaries, one of which is the Affective Norms for English Words [19] and another that has been obtained from [20]. The third dictionary has been derived by combining ANEW with WordNet Synonyms. The weight of a sentence is based on whether it occurs in a verse segment or a chorus segment. The lyric file for each song is segmented into verse and chorus segments prior to calculation of its features. Assigning higher weights to chorus segments has provided better results. This also led to the idea that repeated words have more emotional influence over the user.

To ensure that the proper context of each sentence is taken into account, a set of contextual association rules is used while calculating the AV features. These association rules are based onbasic linguistic principles of the English language and allow modifiers to influence the arousal and valence values of individual words. There are primarily 3 types of modifiers - Verbs, Adjectives, and Negation Words.

Verbs act as modifiers for entire sentences, whereas an adjective acts as a modifier for a specific noun that it is associated with. Negation words can alter the meaning of the verbs and adjectives themselves, which can further affect associated nouns or sentences.

Verbs are very powerful modifiers. A simple example of a verb acting as a modifier could be – "Kill the happy child". The verb "kill" has an inherently dark meaning, and thus converts the entire sentence to a negative one, irrespective of the high valence associated with "happy child". Adjectives act as local modifiers. A number of association rules, based on proximity and sentence construction, are used to tie an adjective to a particular noun. The rules are considered in serial order and are provided below:

- If 2 adjectives occur together, their effects are combined to act as a single adjective
- If an adjective occurs immediately before a noun, the adjective is associated with that noun
- An adjective is associated with the closest noun that occurs before it, unless that noun is already associated with some adjective
- An adjective is associated with the closest noun that occurs after it. In this case, if 2 adjectives are competing for the same noun, the closer adjective is associated with the noun.

Negation words comprise words like "not", "never", "don't", and "can't", among others. All these words are capable of inverting the meaning portrayed by the verb or adjective they are associated with. For example, the sentence – "The boy was not happy". Even though the word "happy" has a high valence, it is negated by the word preceding it. Rules for negation words are applied before applying any of the other modifier rules.

### 3.2.2. BPM

BPM (Beats per Minute) or tempo is indicative of the speed and intensity of emotion associated with a song. A song with a higher BPM is generally perceived to be faster and more energetic than a song with a lower BPM. This feature is used to supplement the Arousal feature, and is calculated with the help of beat tracking algorithms provided by The Echo Nest, an online Music Intelligence Platform.

### 3.2.3. Mode

In the theory of Western music, mode refers to a type of scale, coupled with a set of characteristic melodic behaviors. Modality refers to the placement of an octave's eight diatonic tones and is broadly classified into two types: Major and Minor. The primary distinction between the major and minor modes is the placement of the mediant, or third. In the major mode, the third is comprised of four semitones whereas in the minor mode, the third is comprised of only three semitones [21]. Traditionally, the minor mode has been attributed to feelings of grief and melancholy whereas the major mode has been attributed to feelings of joy and happiness [22], [23].Henver's[24]study on mood associations and major and minor mode, support the aforementioned attributes of mode.

### 3.2.4. Loudness

Loudness of a song is calculated in decibels by measuring the intensity of the audio wave over the song's duration. It provides a good indication of how an average listener would perceive the song and supplements the Arousal feature. Louder songs tend to be more energetic or aggressive, whereas softer songs tend to make use of softer instruments and portray calmer emotions. Unlike traditional loudness measures, the Echonest analysis models loudness via a human model of listening, instead of directly mapping loudness from the recorded signal. This helps to provide a feature that is highly representational of the perception of the average listener.

headerInternational Journal of Artificial Intelligence & Applications (IJAIA) Vol. 6, No. 3, May 2015

### 3.2.5. Danceability

Danceability provides information about the structure of the beats of a song. This feature is meant to represent how suitable the song is for the average dancer. Generally, songs with a higher danceability have more steady beats without much variance in speed. Beats that are very complex or highly varied tend to have lower danceability values, as the listener will find it harder to dance to these beats. The Echo Nest calculates danceability on a scale of 0 to 1, using a mix of attributes such as beat strength, tempo stability and overall tempo.

### 3.2.6. Energy

The energy of a song is the best indicator of the intensity of emotion and supplements the Arousal value obtained from lyric analysis. Energy is calculated on a linear scale with 0 as the lowest value and 1 as the highest value. The calculation itself is performed by the Echonest server, which decodes the audio data from the mp3 file and analyzes it for information.

## 3.3. Feature Scaling

The k-Nearest Neighbors classification algorithm uses a Euclidean distance as a similarity metric (as mentioned in section 5.1). However, if any of the features has a broad range of values, the distance will be governed by this particular feature. Therefore, the range of all features should be normalized so that each feature contributes in proportion to the final distance. Hence, the min-max scaling algorithm has been used to bring all the features in the range of 0-1. Feature scaling is shown in Equation 2.

$$Normalized(e_i) = \frac{e_i - E_{min}}{E_{max} - E_{min}}$$

Equation 2 : Feature Scaling

## 4. TRAINING DATASET

The training dataset has been prepared using last.fm, a popular social musical discovery website. The website allows users to mark songs using any tags they wish and maintains a large repository of songs covering all possible genres of music. Last.fm provides an API which allows easy querying of their database. Several previous works have made use of last.fm to construct their datasets [14], [17], [25], [26].

A list of 9 broad categories with a set of 18 representative tags were selected by studying various emotion models like Russell's Model and Plutchik's Wheel of Emotions. The most popular tags from Last.fm were also considered while constructing the tags. These categories were deemed to cover a sufficiently wide range of emotions.

Last.fm was then queried to retrieve a set of top 50 songs for each of the tags. Duplicate song entries were discarded. In order to ensure a thorough collection of tags, a list of top 20 tags for the remaining 795 songs were then queried from Last.fm. Each of these tags had a weight associated with them which helped to further manually filter the tags. The tags with low weight were discarded and only the ones with high weight were retained. Tags which did not portray any emotion were also removed. For example tags like "Awesome" and "Cool" were removed. This resulted in a comprehensive collection of relevant and meaningful tags. These tags were then distributed into the 9 classes as shown in Table 1.

Based on the tags obtained, each song could be placed into one or more classes. The number of songs in each class is shown in Table 2.

The training dataset was completed by calculating the features for each of the songs. The features were then scaled using min-max scaling in order to ensure simplicity in classification.

Table 1: Classes and their Tags

| Class | Tags |
|---|---|
| Calm | Slow, soft, mellow, peaceful, calm, serene, relaxed, down-tempo, meditative |
| Energetic | Energetic, upbeat, speed, energy, intense, uptempo, metal |
| Dance | Party, dance, dancing, club |
| Happy | Happy, joy, euphoria, ecstatic, cheerful |
| Sad | Sad, anxiety, fear, gloomy, depressed, depression, depressive, melancholic, miserable, misery |
| Romantic | Love, love songs, affectionate, romantic |
| Seductive | Sensual, seductive, naughty, erotic, sexy |
| Hopeful | Hope, hopeful, inspirational, up-lifting, inspiring, lifting |
| Angry | Angry, anger, rage, aggression, aggressive, hate |

Table 2: Number of Songs in each Class

| Class | Number of Songs |
|---|---|
| Calm | 182 |
| Energetic | 103 |
| Dance | 106 |
| Happy | 82 |
| Sad | 223 |
| Romantic | 238 |
| Seductive | 139 |
| Hopeful | 81 |
| Angry | 133 |

## 5. CLASSIFICATION

### 5.1. k-Nearest Neighbors

The idea behind the k-Nearest Neighbor algorithm is quite straightforward. The algorithm makes use of a similarity function to calculate the similarity between any two sets of data. To classify a new song, the system compares the new song with all other songs in the training dataset. It calculates the similarity between the new song and each other song using the provided similarity function.

The system then finds k training songs that have the highest similarity values with the new song. These k songs are referred to as the k nearest neighbors. It then selects the most frequently occurring classes in this set of neighbors and classifies the new song into these classes. The performance of this algorithm greatly depends on two factors, that is, a suitable similarity function and an appropriate value for the parameter k.

The algorithm used in this paper makes use of a Euclidean distance measure that considers n different features to calculate similarity. Euclidean distance is one of the most common distance metrics, and acts as an inverse similarity function. A pair of songs with a high Euclidean distance are considered to be dissimilar, while songs with distances that are low or approaching zero are considered to be similar. Mathematically, the Euclidean distance consists of the square root of the sum of the square differences between each feature of the song to be classified and the training song that it is being compared with. The formula is given in Equation 3.

$$d = \sqrt{\sum_{i=1}^{n}(x_i - y_i)^2}$$

Equation 3: Euclidean Distance

Another good candidate for a similarity function is the Cosine similarity metric, which measures the angle between two vectors. However, through experiments, it was found that the Euclidean Distance with scaled features provided better results than the Cosine similarity metric.

### 5.2. Cross Validation

To minimize the risk of overfitting and to produce more testing data to determine accuracy, the training dataset is partitioned into 4 different subsets to employ a 4-fold cross-validation. The partitioning algorithm randomly picks songs from each category and assigns them to subsets, ensuring to keep a reasonably similar number of songs from each category in each subset. Testing was performed 4 times, each time using one of the subsets as a testing set and the remaining 3 subsets as training sets. Since each song may belong to multiple categories, the number of songs in each subset may differ. However, the representation of each category is approximately equal. Table 3 provides the number of songs in each subset of the training dataset:

Table 2: Cross Validation

| Subset 1 | Subset 2 | Subset 3 | Subset 4 |
|---|---|---|---|
| 197 | 218 | 216 | 164 |

## 5.3. Classification of a New Song

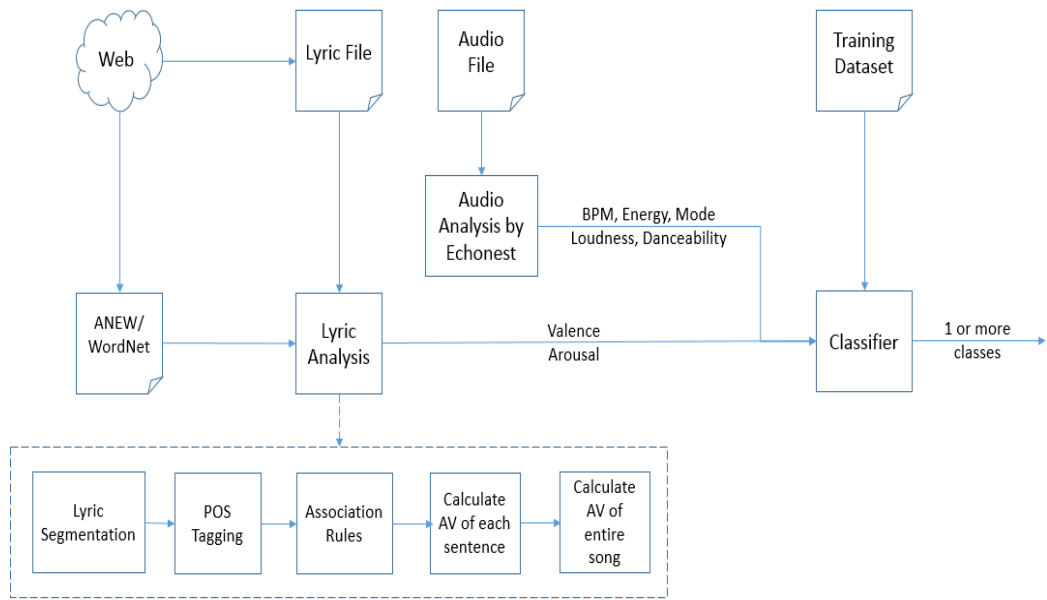

Figure 2: Classification of a New Song

The above figure details the process followed to classify a new song. We rely on the mp3 ID3 tags of a song to fetch the Artist and Title information of the song. The lyric file is then obtained from the Internet by using a lyrics crawler. This crawler will search leading lyric websites to find a match for the Artist and Title and download the associated lyric file for analysis.

Lyric analysis is then performed to calculate the Valence and Arousal features. This analysis follows a 5 step process:

- Segmentation of lyrics into verse and chorus segments
- Perform POS tagging over the entire lyric file
- Apply association rules to each sentence to establish relationships between verbs, adjectives, nouns and negation words
- Calculate the Valence and Arousal of each sentence using the ANEW dictionary along with the association rules
- Calculate the Valence and Arousal of the entire song by using the values of each sentence along with the assigned weights for verse and chorus segments.

These features are then supplemented with audio features that are obtained by analyzing the song on the Echonest servers. All features are passed to the classifier, where they are first scaled and then provided to the kNN classifier. The classifier will classify the song into one or more classes by comparing its features with the training dataset.

## 6. EXPERIMENTS

### 6.1. Feature Weighting

The k-Nearest Neighbors algorithm makes use of a Euclidean distance measure to find the K closest neighbors for a song. However, the standard Euclidean distance formula does not deal with the fact that different features can have varying importance. Some features such as Energy and Valence were found to better characterize a song as compared to other features such as Tempo (Beats per Minute) or Mode. Thus, to improve the classifier, the distance formula was modified to incorporate weights for each feature, and is shown in Equation 4.

$$d_{(x,y)} = \sqrt{\sum_{j=1}^{j} w_j (x_j - y_j)^2}$$

Equation 4: Feature Weighting

Feature Weighting can radically change the list of K nearest neighbors that are returned by the algorithm. Extensive testing with different weights for different features led to a combination that provides a slight increase in accuracy across all test sets. This resulted in an overall increase in accuracy from 82.09% to 83.10%. The formula used is shown in Equation 5.

$$\begin{aligned}
distance = \ &1.0 \times (test_{danceability} - train_{(i)danceability})^2 \\
&+ 0.7 \times (test_{loudness} - train_{(i)loudness})^2 \\
&+ 1.0 \times (test_{valence} - train_{(i)valence})^2 \\
&+ 0.8 \times (test_{bpm} - train_{(i)bpm})^2 \\
&+ 1.0 \times (test_{energy} - train_{(i)energy})^2 \\
&+ 0.5 \times (test_{mode} - train_{(i)mode})^2 \\
&+ 0.9 \times (test_{arousal} - train_{(i)arousal})^2
\end{aligned}$$

where

$train_i$ is the $i^{th}$ training tuple

Equation 5: Weighted Distance Calculation

## 6.2. Fuzzy Classification using a Threshold

Songs cannot usually be classified into just one class. To incorporate this fact, the concept of a fuzzy classification is introduced. In this scheme, a song could be classified into any number of classes, provided that the song has a sufficient number of neighbors in each of those classes. This sufficient number of songs is referred to as a threshold, and its value is crucial to maintaining accuracy with the KNN algorithm.

More formally, the threshold signifies the minimum number of occurrences that a class must have in the song's neighborhood in order to be picked as a potential class for the song under consideration. Given a threshold of 13 and a 'k' value of 30, implies that a class must occur at least 13 times in the 30 nearest neighbors of the song to be a potential class.

There are 2 challenges when using a threshold:

1. Although increasing the value of the threshold increases the accuracy of song classification, it also reduces the fuzziness of the classification
2. In the worst case, a song may not be classified into any of the categories since none of the categories in its neighborhood may have satisfied the threshold condition.

Dealing with these challenges involved extensive testing to select a threshold value that provides a high level of accuracy without significantly compromising the fuzzy nature of the classification. The basic algorithm is as follows:

1. If a song has one or more classes that satisfy the initial threshold condition, classify the song into each of those classes.
2. If there are no classes that satisfy the initial threshold condition, employ the concept of step-wise threshold reduction. In stepwise threshold reduction, the threshold is iteratively reduced until the song is classified into one or more classes. This ensures that all songs are classified into at least one class. Once a song is assigned to at least one class, the algorithm stops reducing the threshold to prevent the introduction of incorrect classes.

## 6.3. Deciding the Number of Neighbors

The number of neighbors (value of K) considered is also important. If the neighborhood is too large, it may introduce several classes that do not match the correct class of the song and also give an advantage to larger classes. On the other hand if the neighborhood is too small, there may not be an appropriate representation of one or more correct classes. The value of k and the value of threshold were chosen to be 30 and 13 respectively. Table 4 compares the performance of the algorithm with and without usage of a threshold.

Table 4 : Using Threshold during Classification

| Set | Without Threshold (%) | With Threshold (%) |
|---|---|---|
| Set 1 | 72.08 | 84.26 |
| Set 2 | 72.61 | 83.49 |
| Set 3 | 69.91 | 82.41 |
| Set 4 | 75.61 | 83.54 |
| Total | 73.46 | 83.40 |

### 6.4. Expanding ANEW with WordNet

In order to expand the domain of words that are considered during the calculation of the Arousal and Valence features, a third dictionary was constructed by combining the ANEW dictionary [19] with WordNet [14], [15], [17], [25], [26], [27]. WordNet is a large lexical database for the English language. It groups words that denote the same concept or meaning (synonyms) into unordered sets called synsets. It also links synsets together by means of conceptual relations (example: Super-Subordinate relationship). Using WordNet, the synonyms for each of the words present in the ANEW dictionary was obtained. The synonyms were then given the same arousal and valence values as the original word in ANEW.

### 6.5. Dealing with Ambiguity and Context

One of the biggest challenges with lyrical analysis is dealing with ambiguity. Words can have multiple meanings based on context and the way sentences are structured. Multiple approaches have been taken to try and gather information from context and association to suitably modify the valence and arousal of a word. This paper makes use of a Parts-Of-Speech (POS) tagger provided by the Python Natural Language Toolkit (NLTK) library, to tag all the words in a lyrics file. Several rules have been used determine the context of a sentence by using adjectives, verbs and negation words as modifiers. These rules are outlined in detail in section 3.2.1.

Lyrics that make use of repeated words or anaphoras tend to do so as a form of emphasis. Reducing the impact of repeated words would fail to capture how the average listener would perceive the lyrics of a song. Hence, no special rules have been used to reduce the effect of repeated words.

The issue of polysemy is harder to deal with as the ANEW dictionary does not account for multiple meanings of a word, and each word can have only one valence and arousal value. The emotion of each word in ANEW represents how the average human perceives that word in isolation, and does not have enough information to distinguish between multiple meanings of a word. To address the issue, a new dictionary would need to be constructed by conducting a survey with sufficient test subjects, similar to the methods used to construct a Chinese version of ANEW (ANCW) in [15]. The improved dictionary can be used in conjunction with Parts-Of-Speech tagging algorithms and the Lesk Word Sense Disambiguation algorithm to properly deal with the issue of polysemy. However, this is currently out of the scope of this paper.

## 7. RESULTS AND CONCLUSION

### 7.1. Observations

Accuracy can be measured by comparing the classes assigned by the algorithm to the classes derived from social tags obtained from last.fm. However, emotions by nature tend to be highly subjective, making it somewhat difficult to identify precisely what constitutes a "correct" classification. Further, since the algorithm employs fuzzy classification, it assigns multiple classes to each song. Some of these assigned classes may not be present in the social tags for that song, but are nonetheless correct. Thus, a good way to identify accuracy is to first identify songs that have been incorrectly classified. The simplest way to do this is to search for any conflicts between the assigned classes and the social tags. A single conflicting tag is sufficient for a classification to be marked as incorrect. A list of conflicting tags is provided in Table 5.

Table 5: Conflicting Tags

| Class | Tags |
|---|---|
| Calm | Energetic, dance |
| Energetic | Calm |
| Dance | Sad, calm, angry, hopeful |
| Happy | Sad, angry |
| Sad | Happy, dance, seductive |
| Romantic | Angry |
| Seductive | Sad, angry |
| Hopeful | Dance, angry |
| Angry | Happy, romantic, angry |

After identifying incorrect classifications, calculating accuracy is straightforward. Results are presented in Table 6.

Table 6: Accuracy

| Set Number | Total Songs | Incorrect | Correct | Accuracy (%) |
|---|---|---|---|---|
| 1 | 197 | 31 | 166 | 84.26 |
| 2 | 218 | 36 | 182 | 83.49 |
| 3 | 216 | 38 | 178 | 82.41 |
| 4 | 164 | 27 | 137 | 83.54 |
| All | 795 | 132 | 663 | 83.40 |

## 7.2. Conclusion and Future Work

In this paper, songs are classified into emotion categories based on a weighted combination of extracted lyrical and audio features. This combination is achieved by using numerous statistical experiments that led to an eventual convergence to the desired weights for each feature. The experiments have obtained a striking balance between accuracy and fuzziness by adjusting the value of threshold and the number of neighbors considered by the kNN algorithm. This classification algorithm can be used to build Music Recommendation Systems and Automated Playlist Generation Systems based on the mood of the user. This research can also be used for analysis of Poems as they are similar to song lyrics.

For further research in this domain, we plan to include more spectral and stylistic features and look further into dealing with the problems posed by word sense ambiguity. We also intend to perform similar techniques on songs in different languages, beginning with Hindi Songs.

**ACKNOWLEDGEMENTS**

The authors would like to thank Prof. Manasi Kulkarni (VJTI) for her help and guidance in this endeavour.

## REFERENCES

[1] Robert Neumayer and Andreas Rauber. Integration of Text and Audio Features for Genre Classification in Music Information Retrieval.Vienna Institute of Technology, Institute of Software Technology and Interactive Systems
[2] Will Archer Arentz, Magnus Lie Hetland, Bjorn Olstad. Retrieving musical information based on rhythm and pitch correlations
[3] Byeong-jun Han, Seungmin Rho Roger B. Dannenberg Eenjun Hwang. SMERS: Music Emotion Recognition using Support Vector Regression.10th International Society for Music Information Retrieval Conference (ISMIR)
[4] Cyril Laurier, Perfecto Herrera. Audio Music Mood Classification Using Support Vector Machine. Music Technology Group, UniversitatPompeuFabra
[5] Patrik N. Juslin and John A. Sloboda. Music and Emotion: Theory and Research. Oxford, UK.
[6] Robert E. Thayer. The Biopsychology of Mood and Arousal. Oxford University Press, 1989.
[7] VandanaKorde, C NamrataMahender. Text Classification and Classifiers Survey.International Journal of Artificial Intelligence & Applications (IJAIA), Vol.3, No.2, March 2012
[8] MaiteTabode, Julian Brooke, Manfred Stede. Genre-based Classification for Sentiment Analysis. Proceedings of SIGDIAL 2009: the 10th Annual Meeting of the Special Interest Group in Discourse and Dialogue, 2009.
[9] Ashis Kumar Mandal and Rikta Sen. Supervised Learning Methods for Bangla Web Document Categorization. International Journal of Artificial Intelligence & Applications (IJAIA), Vol.5, No.5, September 2014.
[10] Menno Van Zaanen, PeiterKanters. Automatic Mood Classification using TF*IDF based on Lyrics. International Society for Music Information Retrieval, 2010.
[11] James A. Russell, Lisa Feldman Barrett. Core Affect, Prototypical Emotional Episodes, and Other things called Emotion: Dissecting the Elephant. Journal of Personality and Social Psychology.
[12] David Rubin and Jennifer M. Talarico. A comparison of Dimensional Models of Emotion: Evidence from Emotions, Prototypical Events, Autobiographical Memories, and Words. US National Library of Medicine, National Institute of Health, PMC, Nov 2010.
[13] Robert Plutchik. The Nature of Emotions. American Scientist, The magazine of Sigma Xi, The Scientific Research Society.
[14] Xiao Hu, J. Stephen Downie. When Lyrics outperform audio for Music Mood Classification: A feature Analysis. 11th International Society for Music Information Retrieval Conference (ISMIR) 2010.
[15] Yajie Hu, Xiaoou Chen and Deshun Yang. Lyrics based Song Emotion Detection with Affective Lexicon and Fuzzy Clustering Method. 10th International Society for Music Information Retrieval Conference(ISMIR) 2009.
[16] Karl F. MacDorman, Stuart Ough and Chin-Chang Ho. Automatic Emotion Prediction of Song Excerpts: Index, Construction, Algorithm Design, and Empirical Comparison.Journal of Music Research, 2007.
[17] Xiao Hu. Improving Music Mood Classification using Lyrics, Audio and Social Tags.University of Illinois at Urbana-Champaign 2010.
[18] Paul Lamere. Social Tagging and Music Information Retrieval.Sun Labs, Sun Microsystems, Burlington, Mass, USA.
[19] Margaret M. Bradley and Peter J. Lang. Affective Normals for English Words(ANEW): Instruction Mangual and Affective Ratings.NIMH Center for the Study of Emotion and Attention 1999.
[20] Amy Beth Warriner, Victor Kuperman and Marc Brysbaert. Norms of valence, arousal, and dominance for 13,915 English lemmas.Behavior Research Methods.
[21] Radocy R.E. and Boyle J.D.Psychological foundation of musical behavior.Springfield, IL 1988.

[22] Lundin R.W.An Objective Psychology of Music.New York, The Ronald Press Company, 1967.
[23] Schoen M.The Psychology of Music: A survey for Teacher and Musician.New York, The Ronald Press Company 1940.
[24] Hevner K.The affective character of the major and minor mode in Music.American Journal of Psychology, 1935.
[25] Xiao Hu and J.StephenDownie. Improving Mood Classification in Music Digital Libraries by combining Lyrics and Audio.University of Illinois at Urbana-Champaign.
[26] Cyril Laurier, Jens Grivolla, Perfecto Herrera. Multimodal Music Mood Classification using Audio and Lyrics.
[27] Dan yang, Won-Sook Lee. Music Emotion Identification from Lyrics.School of Information Technology and Engineering Universitat of Ottawa.

**Authors**

Adit Jamdar has done his B.Tech in Computer Engineering from Veermata Jijabai Technological Institute, Mumbai. He is currently employed as a software development engineer at Amazon Development Centre, Bangalore, India. His interests include Machine Learning, Cryptography and Operating Systems. He is also an avid musician. 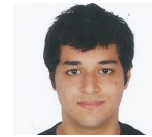

Jessica Abraham has a Bachelor's degree in Computer Engineering from Veermata Jijabai Technological Institute and is currently pursuing a Masters in Computer Science from the University of California at Los Angeles. Artificial Intelligence is her field of interest and she is also a passionate musician. 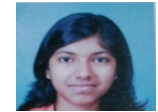

Karishma Khanna has done her BTech in Computer Engineering from Veermata Jijabai Technological Institute, Mumbai. After a summer internship program at Credit Suisse Mumbai, she joined the company as a technology analyst. Her technical areas of interest are Machine Learning and Natural Language Processing. 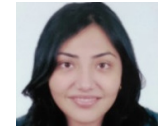

Rahul Dubey has done his B.Tech in Computer Engineering and Technology from Veermata Jijabai Technological Institute, Mumbai. He is currently employed as a software engineer at Samsung Research Institute of Bangalore. His technical areas of interest are Machine Learning, Natural Language Processing and their allied fields. 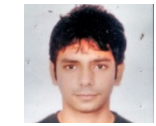